\documentclass[preprint,12pt]{elsarticle}
\biboptions{authoryear}
\usepackage{graphicx}
\usepackage{amssymb}
\usepackage{amsmath}
\usepackage[ruled]{algorithm2e}
\usepackage[utf8]{inputenc}
\usepackage{cleveref}
\journal{Medical Image Analysis}
\usepackage{amssymb}
\usepackage{graphicx}
\usepackage{graphicx}
\usepackage{amsfonts}
\usepackage{amsmath}
\usepackage[table,dvipsnames]{xcolor}
\newcolumntype{L}{@{}>{\kern\tabcolsep}l<{\kern\tabcolsep}}
\usepackage{booktabs}
\usepackage{multirow}
\usepackage{enumitem}
\usepackage[misc]{ifsym}
\usepackage[labelfont=bf]{caption}

\usepackage[table]{xcolor}
\newcolumntype{L}{@{}>{\kern\tabcolsep}l<{\kern\tabcolsep}}

\usepackage{float}  

\usepackage{xpatch,letltxmacro}

\newcolumntype{P}[1]{>{\centering\arraybackslash}p{#1}}

\usepackage{url}
\urldef{\mailsa}\path|{******, *******, *******, *******,|
\urldef{\mailsb}\path|********, *******, *******, *******,|
\urldef{\mailsc}\path|********, *********, *****}*******|

\begin{document}

\begin{frontmatter}

\title{A Collaborative Computer Aided Diagnosis (C-CAD) System with Eye-Tracking, Sparse Attentional Model, and Deep Learning}

\author{Naji Khosravan$^1$, Haydar Celik$^2$, Baris Turkbey$^2$, Elizabeth~C.~Jones$^2$, Bradford~Wood$^2$, Ulas~Bagci$^1$}
 \address{$^1$Center for Research in Computer Vision, University of Central Florida, FL.\\
 $^2$Clinical Center, National Institutes of Health, Bethesda, MD.}


\begin{abstract}
Computer aided diagnosis (CAD) tools help radiologists to reduce diagnostic errors such as missing tumors and misdiagnosis. Vision researchers have been analyzing behaviors of radiologists during screening to understand \textit{how} and \textit{why} they miss tumors or misdiagnose. In this regard, eye-trackers have been instrumental in understanding visual search processes of radiologists. However, most relevant studies in this aspect are not compatible with realistic radiology reading rooms. In this study, we aim to develop a paradigm shifting CAD system, called collaborative CAD (C-CAD), that unifies CAD and eye-tracking systems in realistic radiology room settings. We first developed an eye-tracking interface providing radiologists with a real radiology reading room experience. Second, we propose a novel algorithm that unifies eye-tracking data and a CAD system. Specifically, we present a new graph based clustering and sparsification algorithm to transform eye-tracking data (gaze) into a graph model to interpret gaze patterns quantitatively and qualitatively. The proposed C-CAD collaborates with radiologists via eye-tracking technology and helps them to improve their diagnostic decisions. The C-CAD uses radiologists' search efficiency by processing their gaze patterns. Furthermore, the C-CAD incorporates a deep learning algorithm in a newly designed multi-task learning platform to segment and diagnose suspicious areas simultaneously. The proposed C-CAD system has been tested in a lung cancer screening experiment with multiple radiologists, reading low dose chest CTs. Promising results support the efficiency, accuracy and applicability of the proposed C-CAD system in a real radiology room setting.  We have also shown that our framework is generalizable to more complex applications such as prostate cancer screening with multi-parametric magnetic resonance imaging (mp-MRI).
\end{abstract}


\end{frontmatter}


\section{Introduction}
Lung cancer screening with low dose computed tomography (CT) was shown to reduce lung cancer mortality by $20\%$ (\cite{CAAC:CAAC21387}). Yet, human error remains a significant problem to detect abnormalities. For instance, Missing a tumor (recognition error) and misdiagnosing (decision making error) are called perceptual errors (\cite{kundel1978visual}). It's reported that $35\%$ of lung nodules are typically missed during the screening process (\cite{caroline2014lung}). Over-diagnosis is another significant bias leading to unnecessary treatment which can cause harm and unnecessary medical expenses. 

To alleviate some of these errors, Computer Aided Detection/Diagnosis (CAD) systems have been developed (\cite{firmino2014computer,lemaitre2015computer,jalalian2013computer}). CADs are often known as second opinion tools and they help to reduce false negative findings (i.e., missing tumors by radiologists). CADs also have serious limitations such as a large number of false positive findings and high execution times. Radiologists are expected to eliminate false positive findings generated by the CAD systems, which makes majority of CADs infeasible in routine practice.

%
%
%

Vision scientists have focused on exploring human errors in screening for more than three decades (\cite{kundel1978visual,lee2013cognitive,mccreadie2009eight,al2017eye,kok2017before,venjakob2015visual,drew2013informatics,manning2006radiologists,littlefair2017does,tourassi2013investigating}). One way to explore these perceptual errors is to use eye-tracking technology. It provides information about image interpretation by modeling perceptual and cognitive processes of radiologists. In this paper, we introduce a paradigm-shift system that uses eye-tracking technology as a collaborative tool between radiologists and CAD systems. The rationale behind this idea comes from the fact that radiologists are good at eliminating false positives, which CADs often fail to achieve at a human level performance. On the other hand, CADs capture missing tumors better than the human observer. Because of this complementary properties, we call the proposed technology collaborative CAD (C-CAD). 

Briefly, we develop an accurate and efficient deep learning algorithm that accepts input from an eye-tracking device, and presents the results to radiologists for their diagnostic evaluations. We hypothesize that combining the strength of radiologists and CAD systems will improve the screening/diagnosis performance. To test this hypothesis, we have conduct a lung cancer diagnosis experiment (with low dose CT scans) based on eye-tracking data recorded from multiple radiologists. We also show the applicability of our framework to multi-parametric MRI to conduct prostate cancer screening. We choose lung and prostate cancers due to their high rate of mortality and growth in 2017, thus confirming their clinical importance~(\cite{CAAC:CAAC21387}). 

\subsection{Related Works.}
To the best of our knowledge, the proposed study is the first approach that combines eye-tracking, graph sparsification, and multi-task image analysis in a single framework to make a CAD system. Most relevant studies as compared to individual steps of our algorithms are summarized in the following. Since we focus on two clinical screening examinations in this study (lung and prostate cancers), related studies are confined to only these two topics. Nevertheless, the method presented here can be generalized to other radiology screening examinations.

\textbf{Benefits of Screening:}
According to the American Cancer Society, lung and prostate cancers are the leading causes of death and also the fastest growing cancers in 2017 (\cite{CAAC:CAAC21387}). Medical imaging helps early detection of cancers, but in a recent lung cancer screening clinical trial, it was found that approximately $35\%$ of lung nodules were missed during the screening process by radiologists (\cite{caroline2014lung}). Previous studies have shown that early diagnosis of cancers may have a greater impact on the population~(\cite{CAAC:CAAC21387,caroline2014lung,armato2002radiology}). However, many open questions remain in screening examinations. For instance, definition of at-risk population affects the patient's inclusion in the study. Timing and intervals of screening are adjusted by clinical trials, but there is no optimal method yet to justify these selections. Nevertheless, even in these suboptimal conditions, exciting research is ongoing in this field, thanks to more advanced CT scanners and development of computerized image analysis methods. CAD systems have shown to be useful in reducing  false negative (missed tumor) cases, but the main issue with all CAD systems is the presence of a high false positive rate~(\cite{al2017review}). For instance, when an automated lung nodule detection method was used in a study by ~\cite{armato2002radiology}, 84\% of the missed lung cancers were marked by the computer. Despite this catch, the false-positive rate was very high: 28 false positive findings per scan. 

\textbf{Review of CAD systems in the deep learning era:}
In recent years, deep learning based algorithms revolutionized many fields including medical image analysis applications. In conventional CAD systems (i.e., prior to the deep learning era), hand-crafted feature design/extraction followed by a feature selection and classification scheme were the main steps. However, with the success of deep learning, this strategy has moved from \textit{feature engineering} to \textit{feature learning}. In very recent frameworks, Convolutional Neural Networks (CNN) have been used for feature extraction and off-the-shelf classification methods in most CAD systems~(\cite{van2015off,ciompi2015automatic,tsehay2017biopsy,le2017automated}). In this line of research, for instance, Hua et al. proposed using a Deep Belief Network and a CNN for lung nodule classification~(\cite{hua2015computer}) while Kumar et al. used deep features extracted from an autoencoder to classify nodules into malignant and benign~(\cite{kumar2015lung}). Deep learning based lung cancer detection has also been used as part of a screening strategy~(\cite{roth2016improving,setio2016pulmonary}). 

In our previous works, we have developed various deep learning networks for lung cancer diagnosis~(\cite{buty2016characterization,hussein2017risk,hussein2017tumornet}). In those works, we have first incorporated shape information of lung nodules to improve diagnostic accuracy~(\cite{buty2016characterization}). In another approach, we investigated Gaussian Process algorithms along with CNN to incorporate radiographical interpretations of nodule appearances to improve diagnostic decisions~(\cite{hussein2017tumornet}). Later, we improved our network (called \textit{TumorNET}) by converting the CNN into a multi-task deep learning strategy~(\cite{hussein2017risk}). A multi-task 3D network for joint segmentation and false positive reduction of lung nodules in a semi-supervised manner was proposed in~(\cite{khosravan2018semi}). Meanwhile, many studies in the recent literature focused on false positive reduction in lung nodule detection. Some utilized multiple CNNs for multi-view lung nodule analysis~(\cite{setio2016pulmonary}), while some used 3D CNNs for a more efficient analysis~(\cite{huang2017lung,ding2017accurate}). Furthermore, a multi-scale analysis of lung nodules using multiple 3D CNNs was proposed by~(\cite{dou2017multilevel}). The literature pertaining to lung nodule detection and characterization via CNN is vast. A brief overview of some network architectures related to lung cancer diagnosis can be found in~(\cite{shin2016deep,setio2017validation}). Specific to prostate cancer detection from radiology scans, recent works investigated the application of CNNs using multi-parametric MRI~(\cite{tsehay2017convolutional}) and a semi-supervised approach for biopsy-guided cancer detection using a deep CNN (\cite{tsehay2017biopsy}). Deep learning has also been used extensively as a feature learning tool for various applications such as MRI based prostate segmentation (\cite{guo2016deformable}).

It is beyond the scope of this study to enlist all of the relevant papers in deep learning-based CAD systems in lung and prostate cancers. Herein, we devise a new approach for a CAD design, where deep learning is the part of a collaborative learning strategy. Our proposed framework is generic and its components can be replaced with newer networks, in the future, if desirable.

\textbf{Review of Visual Search Studies in Radiology:}
A key aspect of biological vision studies is to understand perceptual and/or cognitive errors and how radiologists search radiology scans for finding abnormalities. These studies extensively benefit from different eye-tracking technologies~(\cite{venjakob2016review}). Comparison of the visual search patterns of radiologists, and inferring local and global information from those patterns have accelerated the research in this field and led to a better understanding of the differences between expert and novice readers/radiologists, and general strategies for visual search in radiology scans. Some of these studies date back to the 1960s. In spite of decades of work, available methods in the literature fail to provide:
\begin{itemize}
\item A real radiology room experience for radiologists.
\item A quantitative modeling and comparison of eye-tracking data.
\item Exploration of eye-tracking tools’ potential to compensate for CAD errors.
\end{itemize}
In particular, the interaction between radiologists and computers (either simple PACS or CAD systems) remain untouched except by a few seminal image analysis studies~(\cite{drew2013scanners,khosravan2016gaze2segment,venjakob2016image}).

\textbf{Challenges:} Realistic radiology experience with eye-tracking is not achieved yet, mainly because of technical complexities. Eye-tracking systems record data in 2D and having a 3D system needs an exact synchronization. Furthermore, one of the main challenges of quantitative modeling and comparison of gaze data stems from the difficulty of representing, analyzing, and interpreting dense eye-tracking data. This is not only technically challenging, but also computationally demanding~(\cite{venjakob2016review}). The closest study addressing these problems was conducted by (\cite{drew2013scanners}) who proposed the famous \textit{scanning and drilling} paper analyzing gaze patterns at the global level. While \textit{scanners} examine one slice of a radiology scan, more explicitly before moving to the next one, \textit{drillers} keep going forward and backward in slices, moving through a 3D stack of the scan. \textbf{However, quantitative analysis of search patterns at the global and local level has never been addressed before}. We believe that such mapping will be extremely useful since it can serve as a natural interface between radiologists and CAD systems in that details of the gaze patterns will be used to guide a CAD system when it is necessary. In other words, it is not possible to benefit from human visual search in CAD systems when search patterns are represented only at the global level. This forms the major challenge of the problem depicted in (3).

\begin{figure}[h]
\centering
\includegraphics[scale=0.57]{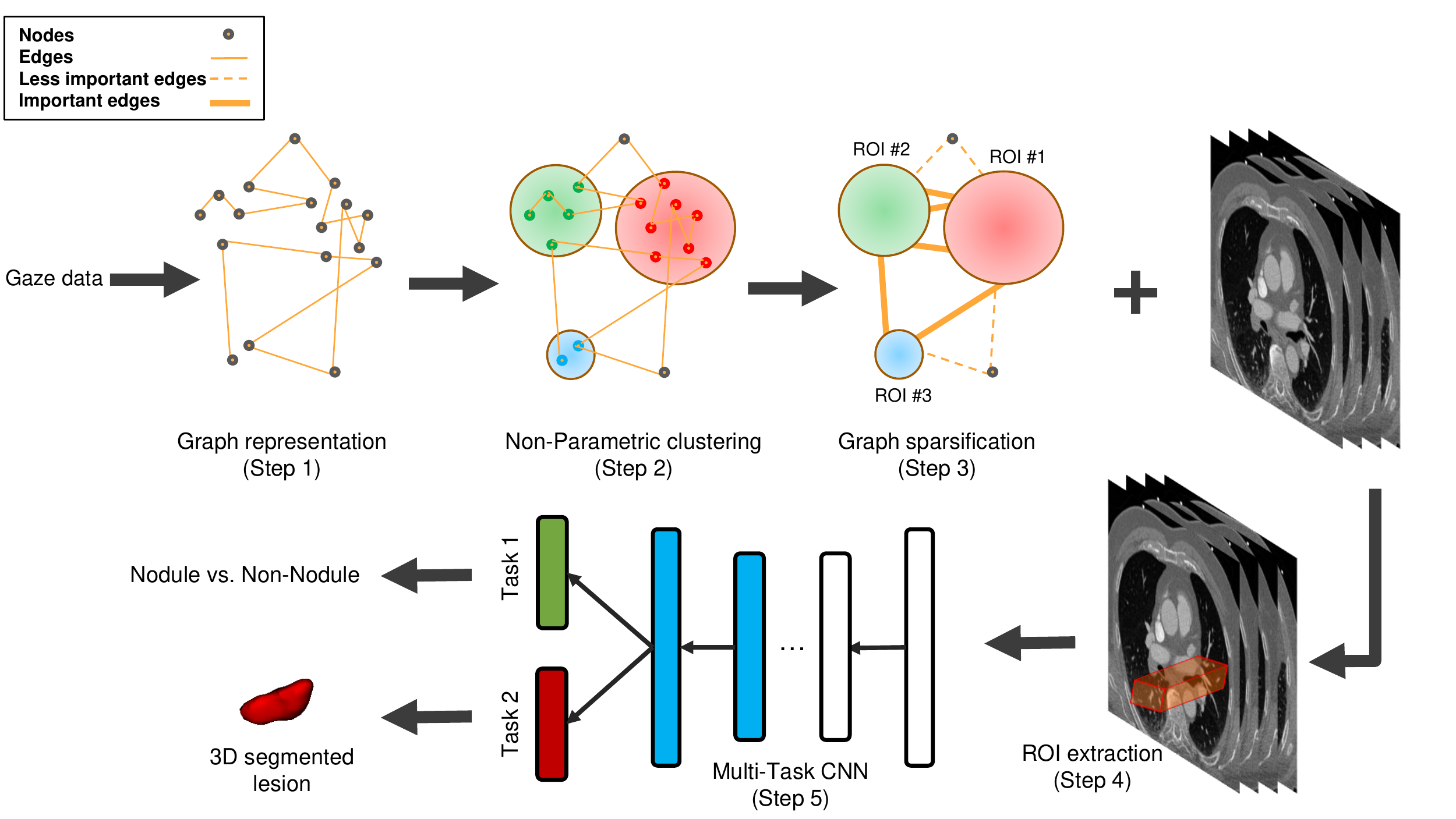}
\caption{To extract radiologist ROIs the dense eye-tracking data goes through a clustering and sparsification algorithm. After extracting ROIs a 3D multi-task CNN is used to perform FP removal and segmentation of lesions inside the ROIs, jointly.\label{fig:overview}}
\end{figure}

\textbf{What we propose?} In this study, we address these challenges by (1) developing an eye-tracking interface that provides a real radiology reading room experience and (2) performing an attention based clustering and sparsification of dense eye-tracking data for building a C-CAD. Our proposed algorithm preserves topological properties of the gaze data while reducing its size significantly. This allows us to quantitatively compare global search patterns of radiologists, extract radiologists' regions of interest (ROI) based on their level of attention during the screening process (local), and to combine this information with image content to do different image analysis tasks for each ROI. Radiologist’s gaze data is represented as a graph and sparsified using the proposed attention based algorithm. Finally, a 3D Deep Multi-Task CNN is presented to perform a joint process of false positive removal (FP removal) and segmentation for each ROI. An overview of the proposed framework (C-CAD) is illustrated in Fig.~\ref{fig:overview} for a lung cancer diagnosis example. Details of each module in the proposed framework are explained in the following sections.  

Our work is built upon our previous effort~(\cite{khosravan2016gaze2segment}), wherein we first thresholded the eye-tracking data by its time component to define potential attentional regions. Then, these regions (ROIs) were processed with computer vision based saliency models to remove some of the false positive regions from considerations. Final ROIs were used for image analysis, particularly for segmenting lung pathologies using attention. A 2D random walk algorithm~(\cite{grady2006random}) was utilized to segment those ROIs by combining visual saliency and visual attention information as seeds of the random walk algorithm. The average dice similarity of $86\%$ was achieved. In the current study, we significantly improve our design into a new level with multiple novel contributions.
\subsection{Our Contributions:}
We believe that our work has significant broader impacts in radiology and imaging sciences and introduces several technical innovations as summarized below:
\begin{itemize}
\item A key aspect of any interactive system is to provide a natural feeling to the user. Having this natural feeling plays a crucial role specifically in the field of radiology, as imposing any constraint might affect the diagnosis accuracy. The majority of eye-tracking studies are conducted in the laboratory settings and no realistic eye-tracking based 3D screening experiment is reported in the literature. Our work fills this research gap and provides a natural (realistic) interaction framework. 
\item We propose a new attention based data sparsification method applied to gaze patterns of radiologists. The proposed approach allows local and global analysis of visual search patterns based on visual attention concepts. More importantly, sparse representation of gaze patterns help interaction with the newly designed CAD system. Our system truly collaborates with its human counterpart (i.e., radiologists); therefore, it is fundamentally different than currently available second opinion tools. 
\item We develop a new CAD system by proposing the state of the art 3D deep learning algorithm in a newly designed multi-task learning platform where both segmentation and diagnosis tasks are jointly modeled. The proposed system has been tested by radiologists with different years of experience, and robustness of the proposed C-CAD has been demonstrated.
\item We extend the proposed C-CAD into a multi-parametric image analysis framework where users can utilize multiple screens as in prostate screening with multi-parametric MRI. To the best of our knowledge, this is the first study in the literature considering multiple screens in the eye-tracking platform.
\end{itemize}

Rest of the paper is organized as follows: In Section 2,  we describe the proposed hardware and software integration, details of data acquisition parameters, the proposed data representation technique with sparsification, and multi-task learning based deep learning algorithm for tumor diagnosis.  In Section 3, we report validation of the sparsification experiments followed by a lung cancer diagnosis experiment with the C-CAD. In Section 4, we introduce the potential of the proposed C-CAD system to handle multi-parametric images and multi-screen based eye-tracking and image analysis in general. We conclude the paper with a discussion and summary in Section 5.

\section{Methods}
\subsection{Data acquisition and Eye-Tracker:}
\begin{figure}[h]
%
%
%
%
\centering\includegraphics[scale=0.6]{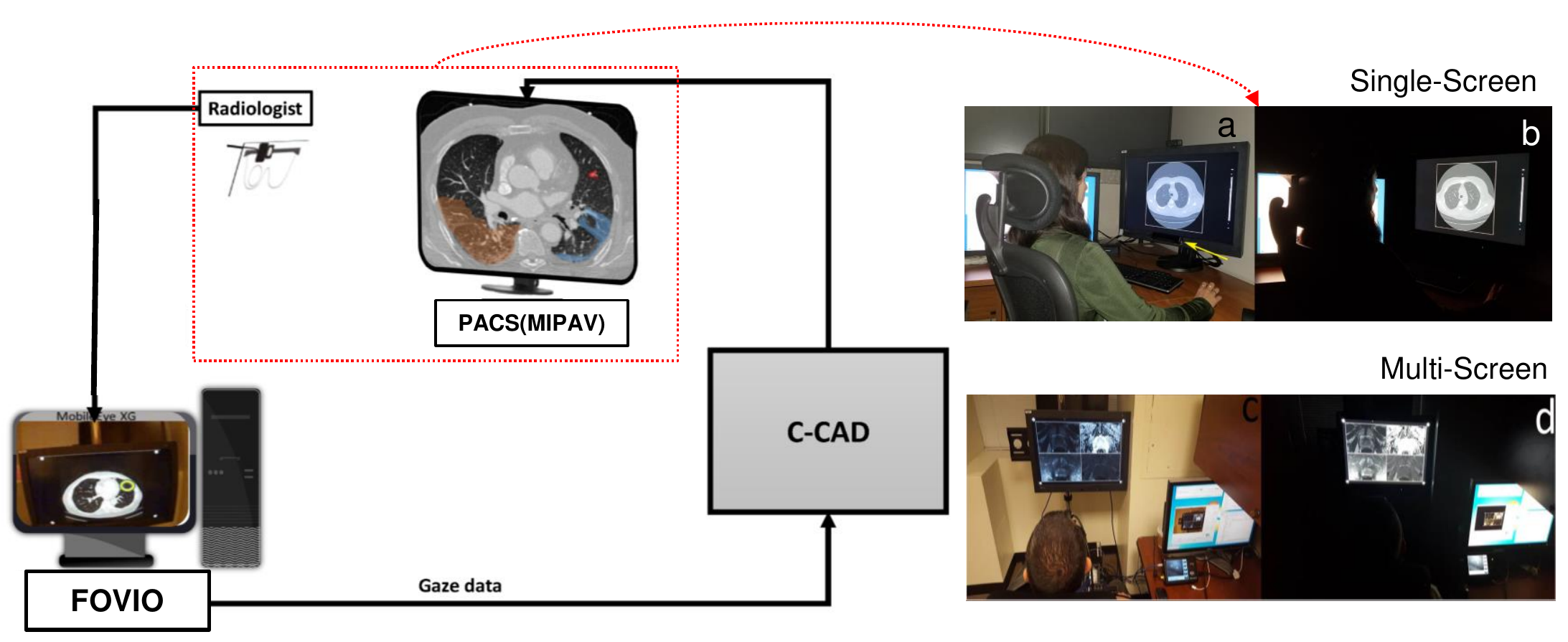}
\caption{A representation of the Eye-Tracking system in a realistic  radiology setting is illustrated. Eye-Tracking system, the connection to the workstation, and the C-CAD system are integrated into the PACS (MIPAV) system directly as shown on the left. Screening experiments in normal light (a,c) and dark (b,d) radiology rooms for single (a,b) and multi-screen (c,d) experiments are shown on the right.\label{fig:data}}
\end{figure}

In this study, a Fovio™ Eye Tracker  remote eye-tracker system (Seeing Machines Inc, Canberra, Australia) was used. We collected eye tracking data using EyeWorks™ Suite ($v. 3.12$) on a DELL Precision $T3600$ using a Windows 7 operating system on an Intel Xeon CPU $E5-1603$ 0 @ $2.80 GHz$ with $8 GB$ of RAM. Figure \ref{fig:data} illustrates integration of Eye-Tracker into our system. Using EyeWorks™, eye movements were recorded by two synchronized, remote eye-trackers at $60 Hz$. All stimuli were presented at a resolution of $1280\times 1024$ on a DELL 19” LCD monitor. We utilized a $60 Hz$ FOVIO eye-tracker and verified calibration through a five-point calibration procedure in EyeWorks™ Record prior to the task. Calibration was considered sufficient if the dot following the eye movement trajectory was sustained (indicating that the eye movement monitor was not losing tracking) and if the calibration dot was accurate (falling on the calibration check targets at the center and corners of the screen when the participant was instructed to look at them, with inaccuracy of up to one centimeter for the upper two corner targets). The eye-tracker was located between $9.5\ cm$ and $8\ cm$ beneath the bottom of the viewing screen (eye-tracker was placed just under the viewing area and at a $25$ degree angle with respect to the monitor.). Following calibration, participants completed the task as described above. After completing this task, the FOVIO was re-calibrated before moving on to a Smooth Pursuit task. Upon completion of screening, the experimental portion of the study was complete and subjects discussed the study with the experimenters before leaving. From consent to debriefing, the study duration spanned roughly $45\ minutes$. Custom made DICOM viewing software was built on Medical Image Processing, Analysis and Visualization software (MIPAV CIT, NIH, Bethesda, MD).

In our proposed framework, eye-tracking data goes through five steps to be converted from a dense graph into a set of diagnostic decisions and segmentation masks on the lesions inside ROIs: \textit{Step 1)} The gaze data is represented with a graph. \textit{Step 2)} A non-parametric clustering method is applied to the graph nodes.\textit{ Step 3)} A novel attention-based sparsification algorithm is applied to the graph to reduce redundant information. \textit{Step 4)} Radiologists' Regions of Interest (ROIs) are extracted based on the level of attention, and  \textit{Step 5)} A deep 3D multi-task Convolutional Neural Network (CNN) is presented to jointly decide about lesions as a nodule or non-nodules (FP removal) and segment potential abnormalities inside ROIs. (See Fig. \ref{fig:overview} for the overview of the proposed system for an example application).

\subsection{Step 1: A graph representation of the eye-tracking data}

\begin{figure}[h]
\centering
\includegraphics[scale=1]{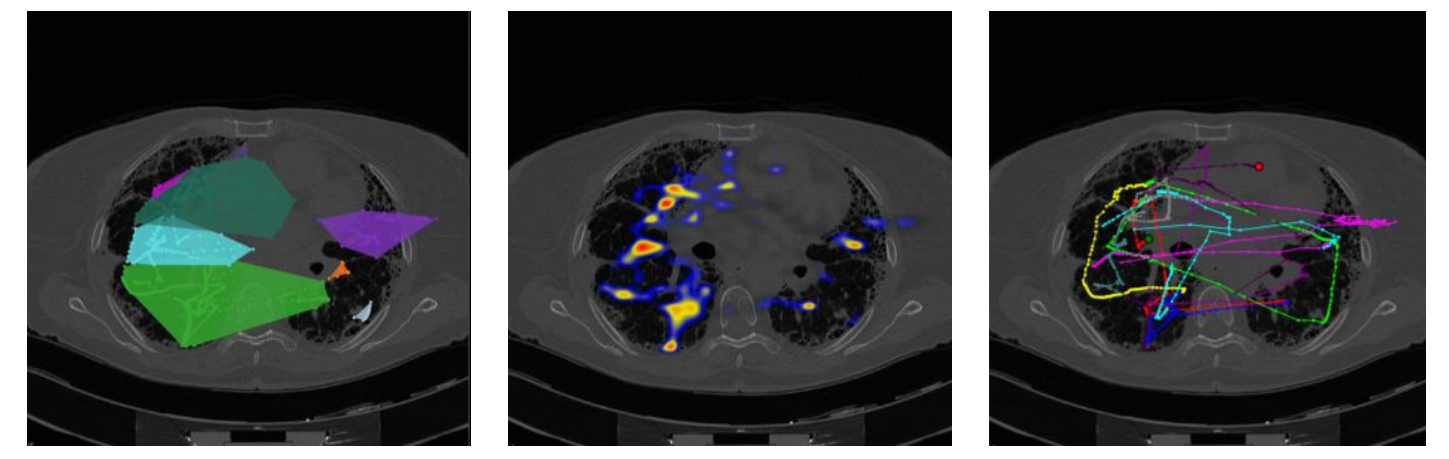}
\caption{Eye-tracking data recorded from lung cancer screening. Low-dose CT is used in a single screen. Gaze patterns (right), heat maps of gaze patterns (middle), and coverage area of the gaze patterns (left) are illustrated. \label{fig:data1}}
\end{figure}

\begin{figure}[h]
\centering
\includegraphics[scale=0.6]{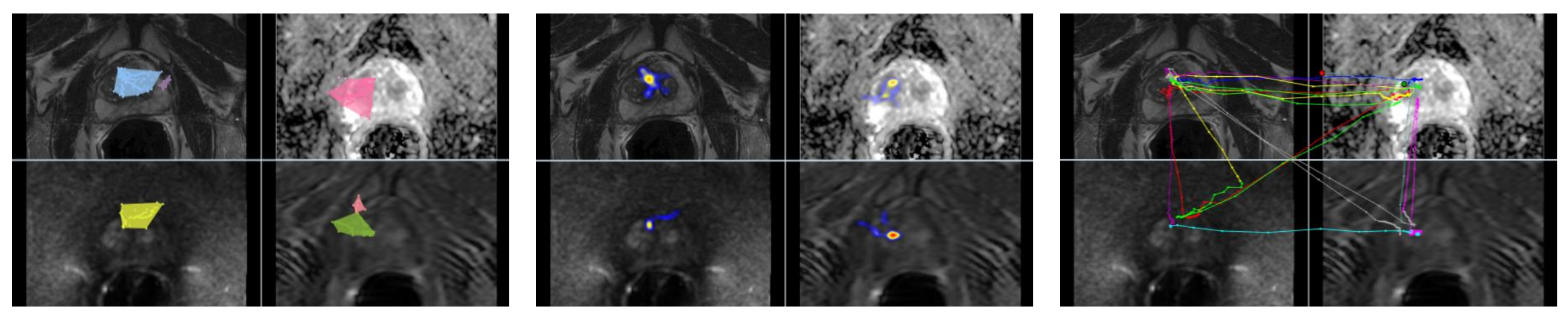}
\caption{Eye-tracking data recorded from prostate cancer screening. Multi-Parametric MRI is used in four screens (left upper: T2-weighted (T2w), right upper: apparent diffusion coefficient (ADC) map, left lower: diffusion weighted imaging (DWI), right lower: dynamic contrast enhanced (DCE) maps). Gaze patterns across different screens and the paths are illustrated for an example screening task. Gaze patterns (right), heat maps of gaze patterns (middle), and coverage area of the gaze patterns (left) are illustrated. \label{fig:data2}}
\end{figure}

\begin{figure}[h]
\centering
\includegraphics[scale=0.425]{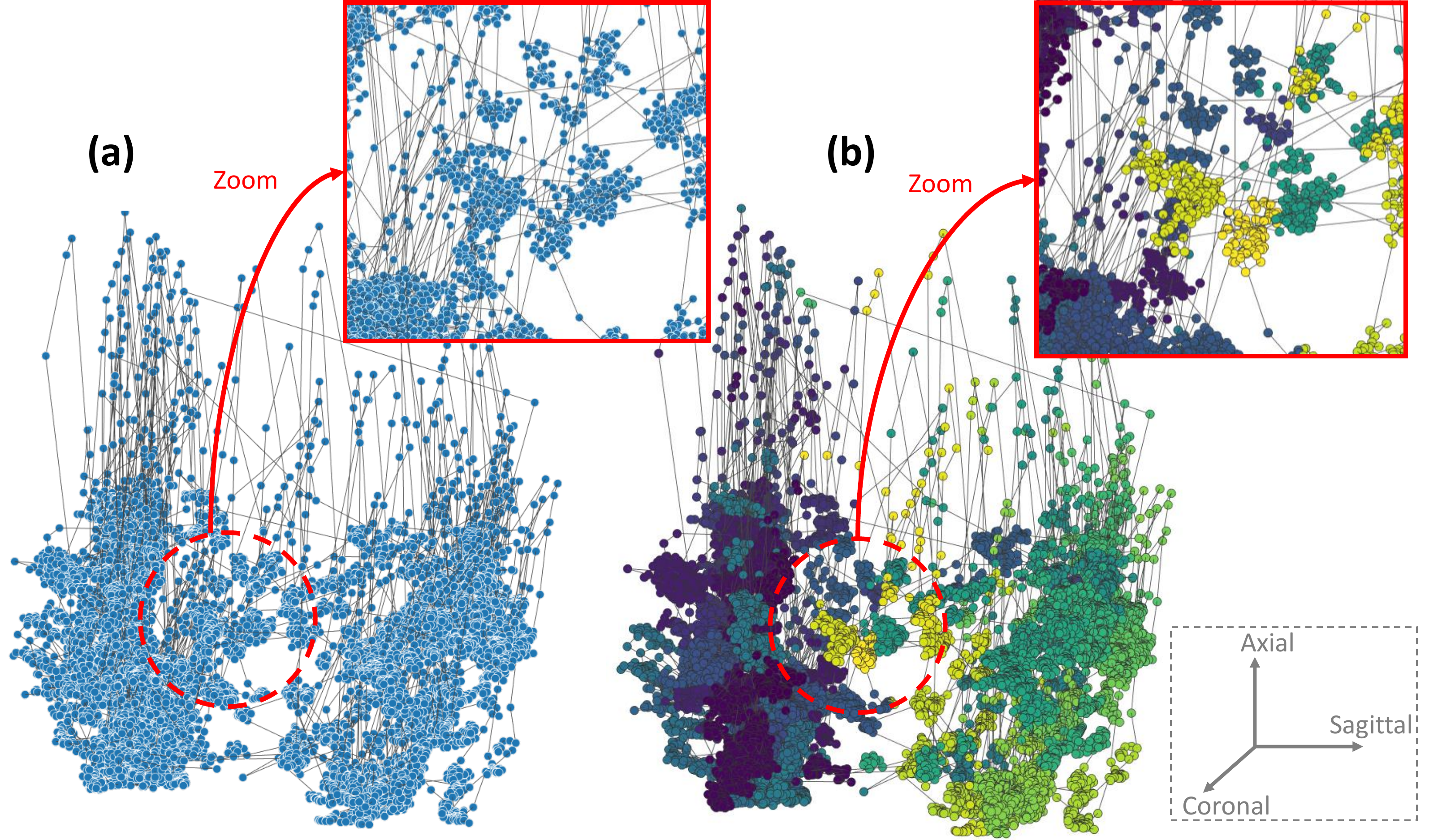}
\caption{(a) 3D Graph representation of visual search patterns from a lung cancer screening experiment. (b) Clustering helps to group gaze points to define attention regions. Colors indicate different clusters.\label{fig:dencevsclustered}}
\end{figure}

An example of eye-tracking data recorded from two cancer screening tasks is shown in Fig.~\ref{fig:data1} and Fig.~\ref{fig:data2}. For each experiment, 2D images are overlaid with the coverage area (left), heatmap (middle), and scanpaths (right) representations inferred from the gaze patterns. Once gaze patterns are recorded, they are dense, hence, difficult to analyze (See Fig.~\ref{fig:dencevsclustered}). The aim in data sparsification is to represent the data with far less parameters and without significantly losing its content. It is also desirable to process the data easily and efficiently when sparisified. To end this, we propose to represent eye-tracking data as a graph and reduce its size without distorting the topology of the data structure by utilizing clustering and sparsification algorithms. 

Graph theory is concerned with a network of points (nodes or vertices) connected by lines (edges). It is a well-established branch of mathematics and it has  numerous successful applications in diverse fields. Formally, a graph $(G)$ refers to a set of vertices $(V)$ and edges $(E)$ that connect the vertices, and it is represented as $G=(V,E)$. In the current problem, a graph representation  is a perfect choice for eye-tracking data because gaze locations (i.e., fixations) can easily be stored as vertices while path/directions (i.e., saccades) between gaze locations can be stored as edges in the graph. An example of a 3D graph representation of gaze patterns obtained from a lung cancer screening experiment using volumetric low-dose CT scans is illustrated in Fig.~\ref{fig:dencevsclustered}(a). A zoomed version of the graph indicates dense data points and edges between them. For simplicity, edges are shown as undirected.

Although graph representation allows parameterization of the patterns in the data, its analysis and interpretation are infeasible because the graph includes large amount of nodes and edges as exemplified in Fig.~\ref{fig:dencevsclustered}. Such graphs are called "dense", and \textit{sparsification} operation can be considered to simplify the data to overcome challenges of the analysis of a dense graph. A graph sparsification algorithm reduces the graph density by omitting unnecessary edges. However, there are challenges unique to our problem which makes the conventional sparsification algorithms suboptimal in our case:
\begin{itemize}
\item \textit{First}, the constructed graph is consecutive, meaning each edge in our graph connects only two distinct vertices (due to the nature of eye-tracking data). This causes a maximum vertex degree of 2 in our graph resulting in the failure of current sparsification algorithms to remove even a single edge from our graph. This is because all the edges in our graph are considered equally important for keeping the structure of the graph unchanged. 
\item \textit{Second}, the radiologist's attention should be taken into account while sparsifying the data. This will  make sure that the global visual search pattern and attention regions are both preserved after sparsification.
\end{itemize}

To overcome these challenges, we propose the next two steps of our algorithm to handle the consecutive nature of the data and encountering attention information, respectively.

\subsection{Step 2: Non-Parametric clustering of the graph nodes}
We propose to apply a non-parametric clustering algorithm to graph vertices of gaze data and reconstruct the graph from clustered vertices (i.e., one vertex for each cluster). There exists a great number of clustering methods in the literature due to its applicability in many fields. Each algorithm has advantages and disadvantages. Since gaze patterns are dense, it is desirable that the clustering algorithm is chosen from a time-efficient family of algorithms.

Non-Parametric clustering algorithms often have subjective measures for partitioning the data into distinct groups. Hence, many efforts are spent on designing such measures. The choice of this similarity measure (or called dissimilarity metric in the same fashion) is very important, as it has a strong effect on the resulting groupings. In the current problem, we make a domain-specific definition of \textit{similarity} for gaze patterns. Simply, we use distance between gaze points as a similarity measure. This is because if radiologists spend more time in screening for a particular region (i.e., attention region), then the data collected from those regions are dense and in close vicinity of each other. Likewise, if the distances between gaze locations are far, it can be safely assumed that a different attention region is being examined. Given the fact that, there will be some heterogeneity in distance measurements, it is still reasonable to use it, as is common in most non-parametric clustering algorithms. However, any clustering algorithm with a distance based similarity metric will not be an optimal choice in our case since the data is extremely dense, and we desire the algorithm to run in linear time while processing large amounts of data. 

We hypothesize that the Balanced Iterative Reducing and Clustering using Hierarchies (BIRCH) algorithm can be a good fit for our purpose because it is time-efficient (linear time), non-parametric, and local~(\cite{zhang1996birch}). BIRCH uses a Cluster Feature $(CF)=(N,LS,SS)$ to make a large clustering problem tractable by concentrating on densely occupied portions. For $N$ data points in a given cluster $X_i$, $LS=\sum_{i=1}^NX_i$, and $SS=\sum_{i=1}^NX^2_i$ are used to measure pairwise distances between data points. $N$, $LS$, and $SS$ are basically representing norms of the cluster.

This step represents the dense data with a set of attention regions by clustering them into different groups allowing us to modify our graph as follows: 
\begin{enumerate}
\item all the vertices pertaining to each cluster are removed except the centroid of the cluster, 
\item all the edges that were connecting vertices from different clusters now connect the corresponding centroids, and
\item all the edges that were connecting vertices inside each cluster are modeled as self loops on the centroid (see Fig.~\ref{fig:overview}).
\end{enumerate}

This modification allows the maximum vertex degree of larger than 2 in the graph. The results of clustering on the graph vertices and the modified graph are shown in  Fig.~\ref{fig:dencevsclustered}.  (Note: the self loops on cluster centroids are not shown in the modified graph for the sake of visibility). Spectral sparsification algorithms can now be used to reduce the data further.

\subsection{Step 3: Attention based graph sparsification}
In the previous step, we simplify our graph by eliminating vertices inside each cluster and represent them by a centroid. This step will switch the attention mechanism from vertices to clusters, meaning that the attention region is now represented by clusters. Graph sparsification methods, on the other hand, eliminate edges and convert a dense graph $G$ into a sparse graph $S$.

In our proposed method, we intend to preserve the structural similarity between the sparsified and original graphs; therefore, spectral sparsification algorithms are more suitable than the other kinds. It is mainly because many graphs can be characterized better with spectral estimations where spectral information is obtained by an adjacency matrix, which is normalized by its edges and subtracted from the identity matrix. In our problem, another constraint is to have a linear or nearly-linear sparsification algorithm so that the whole dense data analysis become efficient. Due to these two constraints (being fast and intention to preserve structural similarities), we adapted Spielmans's \textit{nearly-linear time} spectral graph sparsification method (\cite{spielman2011graph}) with a novel weight parameter, $w$, to reflect the attention regions inferred from eye-tracking data. This particular spectral sparsification algorithm forces the Laplacian quadratic form of the sparsified graph to be a $\sigma-spectral \ approximation$ of the original graph (\cite{spielman2011graph}) and preserves the structural similarity between the sparsified and original graph. Note that the  spectral sparsifier is defined as a subgraph of the original whose Laplacian quadratic form is approximately the same as that of the original graph on all real vector inputs, as proved by (\cite{spielman2011graph}). That is, if the Laplacian (i.e., spectral properties) matrix is preserved, the structural similarity is preserved. 

The Laplacian matrix of a weighted graph $G(V,E,w)$ is defined simply as:

\begin{equation}
L_{G}(i,j)=\left\{\begin{array}{lr}
-w_{i,j}  &  i\neq j\\ 
\sum_{z}w_{i,z}&  i=j,
\end{array} \right.
\label{eqn:Laplacian}
\end{equation}

\noindent where $w_{i,j}$ represents the weight of edge between vertices $i$ and $j$. The Laplacian quadratic form of G for $x \in{\rm I\!R}^{V}$ is:

\begin{equation}
x^{T}L_{G}x=\sum_{i,j\in E}w_{i,j}
(x(i) - x(j))^{2}.
\label{eqn:Laplacian quadratic}
\end{equation}
Let $\hat{G}$ be a $\sigma-spectral \ approximation$ of G if for all $x \in{\rm I\!R}^{V}$ such that:
\begin{equation}
\frac{1}{\sigma}x^{T}L_{\hat{G}}x \leq x^{T}L_{G}x \leq \sigma x^{T}L_{\hat{G}}x \ .
\label{eqn:specteral approximation}
\end{equation}
In the original implementation of the spectral sparsification algorithm (\cite{spielman2011graph}), a weighted graph $G(V,E,w)$ is converted to a sparse graph $S(V,\hat{E},\hat{w})$ with $|\hat{E}|= O(n \log n/\alpha^{2})$ in $\tilde{O}(m)$ time, for a graph with $n$ vertices and $m$ edges, such that

\begin{equation}
\small
(1-\alpha)\sum_{i,j\in E}w_{i,j}
(\delta x)^{2}\leq\sum_{i,j\in \hat{E}}\hat{w}_{i,j}
(\delta x)^{2} \leq (1+\alpha)\sum_{i,j\in E}w_{i,j}
(\delta x)^{2},
\label{eqn:specteral approximation2}
\end{equation}
\noindent where $\alpha$ is the sparsification parameter and $\delta x$ denotes $x(i) - x(j)$. This method samples edges from $G$ with a probability proportional to $w_{i,j}.r_{i,j}$, where $r_{i,j}$ represents effective resistance corresponding to the edge. Note that \textit{effective resistance} is a distance measure, inspired by the homology (i.e., correspondence) between a graph and an electrical network. 

We modify the spectral sparsification approach according to our unique problem by adding a novel weight function to it. We present the radiologists' level of attention on different regions by the edge weight between those regions. These weights are also used as a probability measure to define their importance. More important edges are defined as the edges connecting regions with more attention (indicating dense visual search patterns). We transfer these rationale into our graph using two primary parameters:
\begin{itemize}
\item \textit{Number of nodes in each cluster (N)}: indicating a \textbf{global} representation of attention for a particular region. The more a radiologist spends time on a region, the denser the corresponding cluster is. 
\item \textit{The amount of time spent on one cluster (C)}: indicating a \textbf{local} representation of attention for a cluster. The number of self-loops on each centroid can be considered to define such parameters.
\end{itemize}

We then define edge weights based on these two parameters as below:
\begin{equation}
w_{i,j}=exp({-N_{i}^{2}\times C_{in}})^{-1}\times exp({-N_{j}^{2}\times C_{jn}})^{-1},
\label{eqn:specteral approximation weight}
\end{equation}
\noindent where $N_{i}$ and $N_{j}$ are number of nodes in clusters $i$ and $j$, respectively, and $C_{in}$ and $C_{jn}$ are number of self-loops for clusters $i$ and $j$. Each $exp$ term can be considered as the attention level corresponding to each of the nodes. The major strength of our modified spectral sparsification algorithm is that we model both local and global visual patterns and their interactions through this weight function. The pseudo code of our sparsification method is given in Algorithm \ref{alg:Sparsification}. 
\begin{algorithm}
    \SetKwInOut{Input}{Input}
    \SetKwInOut{Output}{Output}

    \Input{Dense graph: $G=(V,E)$, $\alpha$: Sparsification parameter}
    \Output{Sparse graph: $S=(V,\hat{E},\hat{w})$}          
    \For{$i$ and $j$ in $V$}{
        \If{ There exists $e_{i,j}$ in E}{
    	Compute $N_{i}$ and $N_{j}$ \begin{footnotesize}(number of nodes in clusters corresponding to vertices $i$ and $j$)\end{footnotesize}\\
    Compute $C_{in}$ and $C_{jn}$ \begin{footnotesize}(number of self-loops corresponding to vertices $i$ and $j$)\end{footnotesize}\\
    Compute $w_{i,j}$ using eq.\ref{eqn:specteral approximation weight}\\
    }
      }
return $G(V,E,w)$\; 
\For{$e_{i,j}$ in E}{
Sample edge $e_{i,j}$ form $G$ with prob. proportional to $w_{i,j}.r_{i,j}$\\
Add $e_{i,j}$ to $S$
}
return $S(V,\hat{E},\hat{w})$\; 
    \caption{Attention based Spectral Graph Sparsification}
    \label{alg:Sparsification}
\end{algorithm}

\begin{figure}[h]
\centering
\includegraphics[scale=0.54]{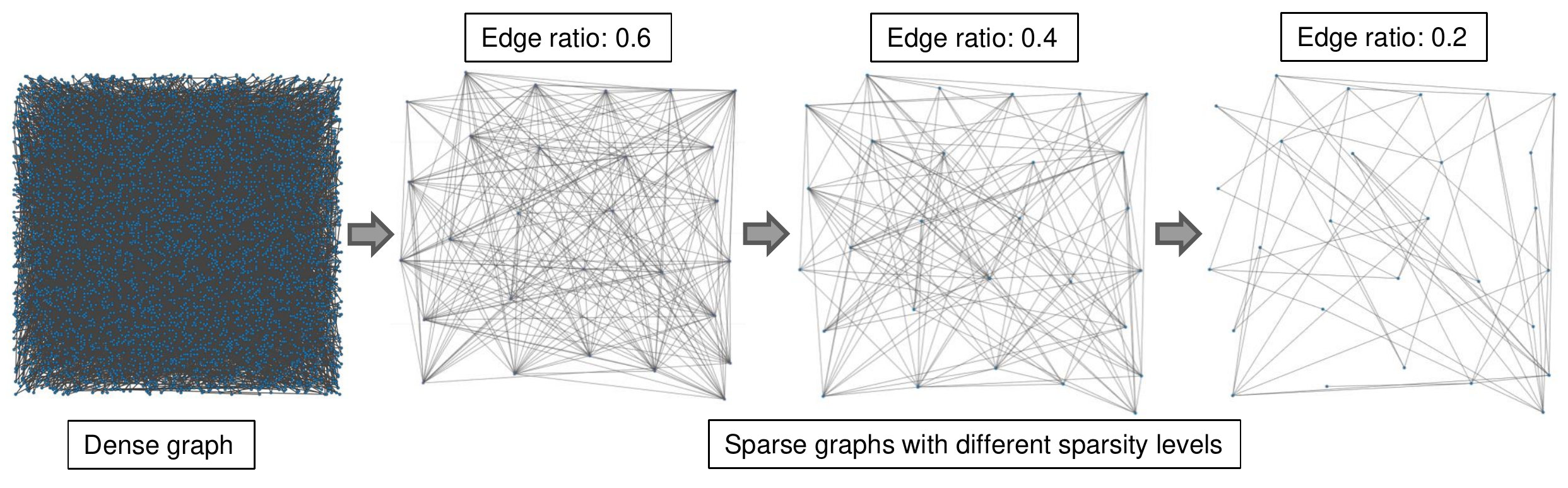}
\caption{Results of applying proposed graph sparsification method on a 2D dense synthetic data. Edge ratio is the ratio of edges after applying the method to the original graph. \label{fig:2Dtoy}}
\end{figure}

To illustrate the effect of our sparsification method on very dense data, we applied the proposed method on synthetic data with different sparsification levels. Result is shown in Fig.~\ref{fig:2Dtoy}. The data was created by randomly generating locations which were connected to each other consecutively to best mimic human gaze (in terms of consecutive connections). Progressively sparsified graphs are shown with respect to different levels of edge ratio. \textit{Edge ratio} herein refers to the percentage of the total number of edges remaining after the sparsification; hence, the most sparse graph is represented on the last column where the edge-ratio is set to $0.2$.

\subsection{Step 4: Extracting attention-based ROIs}
Having discussed how to construct the sparsified graph while keeping attention information, we now discuss how to extract the \textit{Regions of Interest (ROIs)} from the graph. This step allows us to combine this information with image content, all in 3D space, and perform different analysis inside ROIs.

The attention level of each node inside our graph ($a_{i}$) is defined as a combination of global and local attention information on that node. This can be formulated as follows:
\begin{equation}
a_{i}=exp({-N_{i}^{2}\times C_{in}})^{-1},
\end{equation}
\noindent where $N_{i}$ represents the number of nodes in cluster $i$ (i.e., global attention level) and $C_{in}$ represents the number of self loops on cluster $i$ (i.e., local attention level). That is, higher values of $a_{i}$ correspond to higher focus of attention on a corresponding cluster centroid. The cluster centroid represents a location in the 3D space of image coordinates. 

Our method enables us to perform any image analysis tasks on the ROIs including but not limited to segmentation, detection of particular pathology, and diagnosis. In the next section, we demonstrate how the search pattern and attention information from the radiologists' gaze can be combined with image content to perform radiological image analysis: nodule segmentation and false positive removal. 

\subsection{Step 5: A 3D multi-task deep learning for joint nodule segmentation and false positive removal}
We propose a fully 3D deep multi-task CNN that can perform two image analysis tasks simultaneously. Multi-Task Learning (MTL) is a branch of machine learning that deals with jointly optimizing multiple loss functions to update a single model. The goal of MTL is to generalize a learned model to be able to perform multiple tasks~(\cite{caruana1998multitask}). Other than having a generalized model, MTL can benefit each task by learning from other tasks. Some underlying features learned from one task can be helpful in increasing the performance of model on another relevant task. 

In this work, we propose to use a single 3D CNN for combining two relevant tasks of segmentation and false positive (FP) removal. We chose segmentation and FP removal tasks because morphology/volume and shape are important features to distinguish nodules from non-nodules, and decision makers for determining follow up scan time or necessary intervention. Besides, learning a segmentation task in parallel to FP removal task strengthens the effect of such features. The proposed network segments the object inside the ROIs regardless of it being a nodule or not, while classification task scores the presence of nodule inside 3D ROI, which makes these two tasks relevant. In the results section, we show that both tasks benefit from each other and the network performs better on both. This justifies that knowing an abnormality as nodule or non-nodule helps segmentation and the segmentation helps in decision making about nodule vs. non-nodule in return.

For the experiment, we briefly feed a 3D volume of interest (VOI) around the cluster centroid corresponding to radiologist’s ROIs into our 3D deep multi-task CNN to perform FP removal and segmentation jointly. The proposed CNN architecture is an encoder-decoder network containing $14$ shared layers, $6$ of which are convolutional layers in the encoder side, and $6$ convolutional layers are in the decoder side. There are also one down sampling and one up-sampling layers. The shared layers are trained jointly for both tasks while task specific layers are trained only on a single task. The network has $3$ task specific layers including $1$ sigmoid layer, specific to segmentation, and $2$ fully connected layers, specific to false positive (FP) removal task. The fully connected layers have $1024$ and $2$ neurons, respectively.

\begin{figure}[h]
\centering
\includegraphics[scale=0.55]{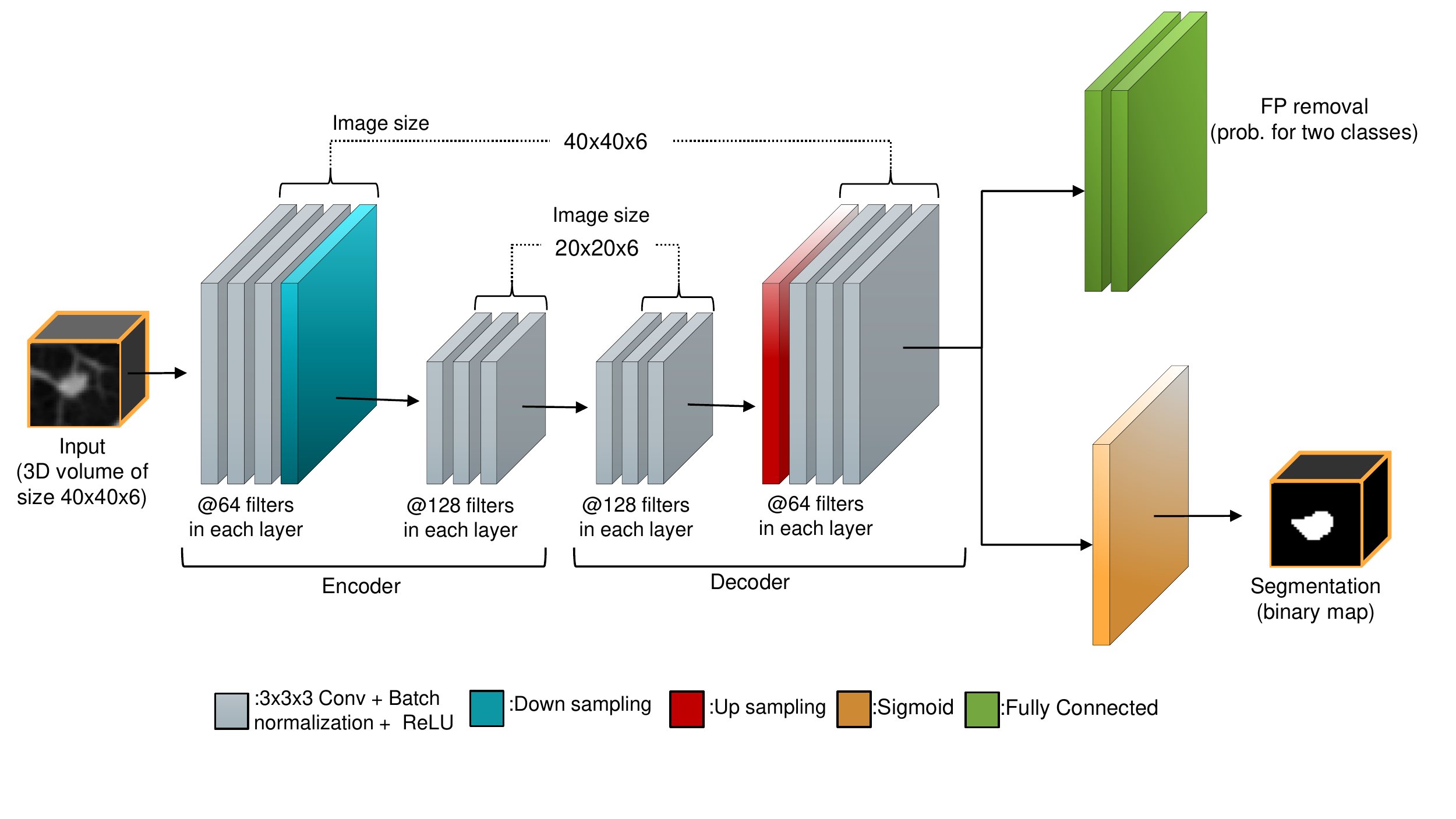}
\caption{The 3D deep multi-task CNN architecture. The size of all convolutions are $3\times 3\times 3$ with a stride of $1$ in each dimension. The downsampling and upsampling operators are performed only in the $xy$ plane, and do not affect temporal information. All convolution and layers are $3D$. The network has $14$ shared layers, $2$ FP removal specific layers and one segmentation specific layer. \label{fig:architecture}}
\end{figure}

The input to the network is a $3D$ volume of size $40\times 40\times 6$ centered at the centroid corresponding to the ROIs. The output for FP removal is the probability of lesion membership to one of the classes (nodule vs. non-nodule) and the output of the segmentation task is a binary mask of a nodule. Our network architecture is inspired by the work by~(\cite{badrinarayanan2017segnet}) and can be considered as a significant extension of their work into a \textit{3D} \textit{multi-task} architecture. The details of our different architecture are summarized in Fig.~\ref{fig:architecture}.

\section{Results}
\subsection{Data}
We tested our proposed method on synthetic data and two real screening experiments:

\underline{Synthetic data:} included random generation of $5000$ nodes with consecutive generated edges between these nodes, to better mimic the nature of eye-tracking data. 

\underline{Lung cancer screening:} the chest CT scans for this experiment were obtained from  Lung Tissue Research Consortium (LTRC) (https://ltrcpublic.com/). The in-plane resolution of the CT images are $512\times 512$ with a voxel size of $0.58\times 0.58\times 1.5~mm^{3}$. For training our MTL based CNN, we used the data provided by LUNA challenge~(\cite{LUNA16}). The challenge organizers provided the location of nodules accepted by at least $3$ out of $4$ radiologists. Scans with a slice thickness greater than $2.5$ mm were excluded from the dataset. Overall, we used $6960$ lung nodule ROIs (in 3D) to train/test our network, $3480$ of which were positive samples (nodules which were the augmentation of $498$ distinct samples) and the rest were negative samples (non-nodule). Positive samples (nodules) were augmented by shifting into $6$ directions (top, bottom, left, right, top-left and bottom-right) to keep the balance between classes. The binary masks were obtained manually by a trained annotator using the ITK-SNAP 3D segmentation tool~(\cite{yushkevich2016itk}). For MTL-based deep learning experiments, $20\%$ of data was used for testing while the rest was used for training, and this procedure was repeated five times to get average score from all experiments.

\underline{Participants:} Three radiologists with different expertise levels participated in our experiments. Their reading experience levels were noted as $20$, $10$, and $3$ years, respectively. After necessary adjustment and calibrations of eye-tracking equipment were done for each participant, they were given instructions about the screening process without letting them know the existence or absence of tumors in the scans. Mouse and other manipulations (zoom, scroll, contrast/brightness window selection) were recorded automatically by the software along with gaze locations.

\subsection{Evaluation of Graph Sparsification}
In order to show the effectiveness of our proposed graph sparsification method, we tested it first on a dense synthetic data. Then, a $3D$ lung cancer screening experiment was performed. To show that our algorithm is capable of analyzing multi-screen experiments, we applied the proposed algorithm on a 3D multi-parametric MRI prostate cancer screening as a feasibility study. All of our experiments were performed in a real radiology room setting without putting any restriction or limitations on the radiologist. The qualitative and quantitative results of the above-mentioned experiments are reported in the following sections.

\subsubsection{Synthetic data experiment:} 
\begin{figure}[h]
\centering
\includegraphics[scale=0.42]{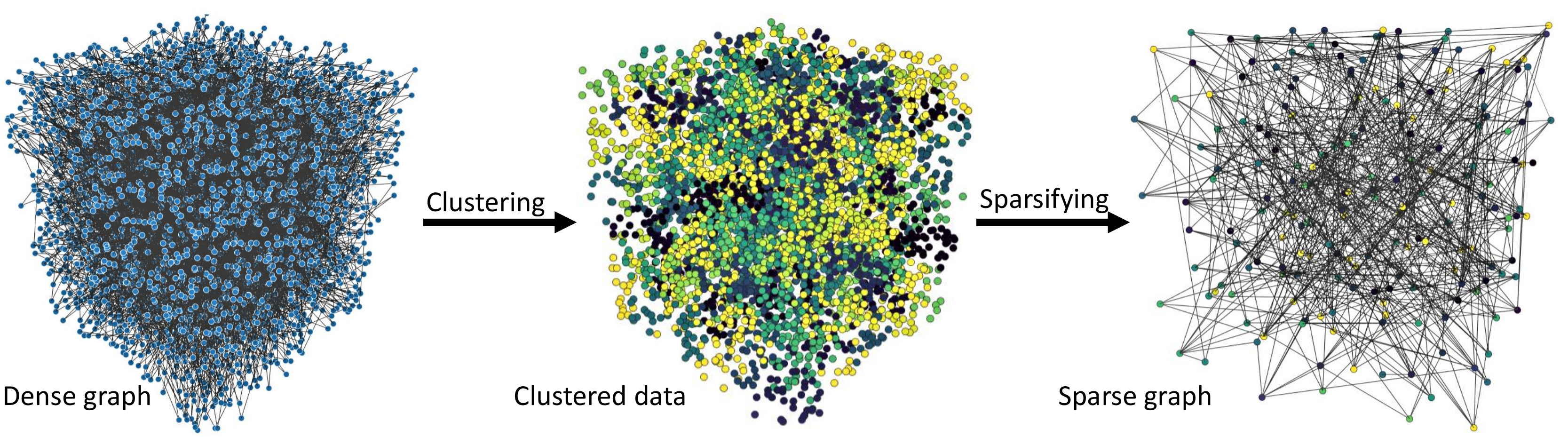}
\caption{Sparsification results from synthetic data experiments. The number of graph nodes are reduced from $5000$ to $196$ in the clustering step, and the number of edges (after clustering) are reduced from $4269$ to $524$ in the graph sparsification step. \label{fig:toy3d}}
\end{figure}

The goal of this experiment was to show the ability of our proposed method in dealing with very dense data. In the synthetic data, $5000$ nodes were connected consecutively in the $3D$ space to create a dense data. The reason behind consecutive connections is to mimic data recorded from eye-trackers. Figure \ref{fig:toy3d} illustrates the effect of our algorithm in sparsifying the data. The number of edges were reduced from 4269 to 524 in sparsification step, and the number of nodes were reduced from 5000 to 196 in the clustering step.

\subsubsection{Lung cancer screening experiment:} 
\begin{figure*}
\centering
\includegraphics[scale=0.40]{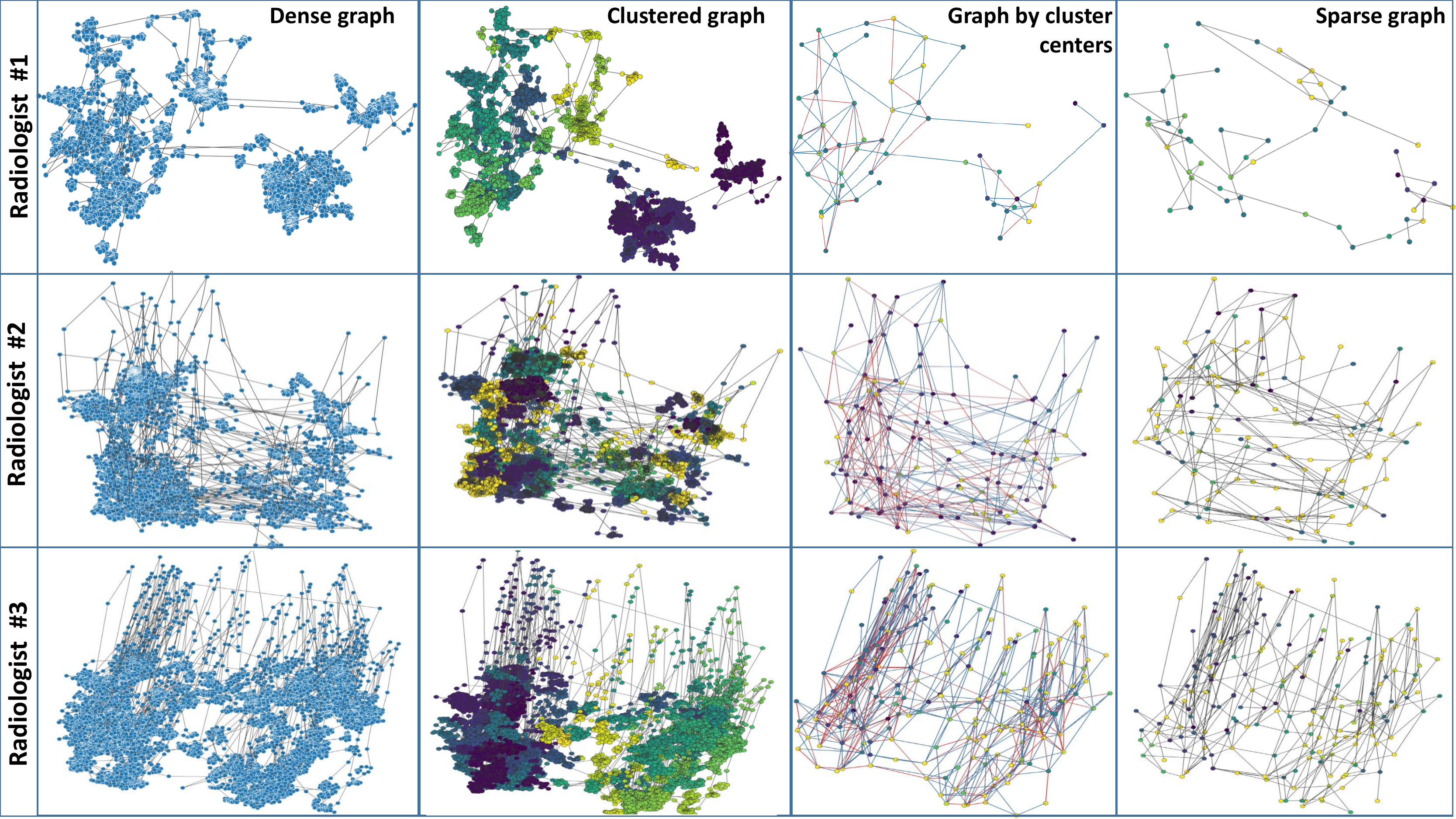}
\caption{Lung cancer screening experiments with CT data. First column: dense gaze patterns. Second column: attention based clustering. Third column: nodes in clusters are reduced. Fourth column: sparse graph after further reducing edges. \label{fig:qualitativelung}}
\end{figure*}

In the lung cancer screening experiments, our participants examined volumetric chest CT scans and the corresponding data was recorded in 3D space. The qualitative results and comparison of visual search patterns of the three radiologists are reported in Fig.~\ref{fig:qualitativelung}. Each column shows one step of the proposed algorithm and each row corresponds to a radiologist. As can be seen, dense graphs hardly reveal any comparisons between radiologists' visual search patterns. However, it is much easier to use sparsified graph (last column) for a global comparison of visual search pattern. 

\begin{figure}[h]
\includegraphics[scale=0.59]{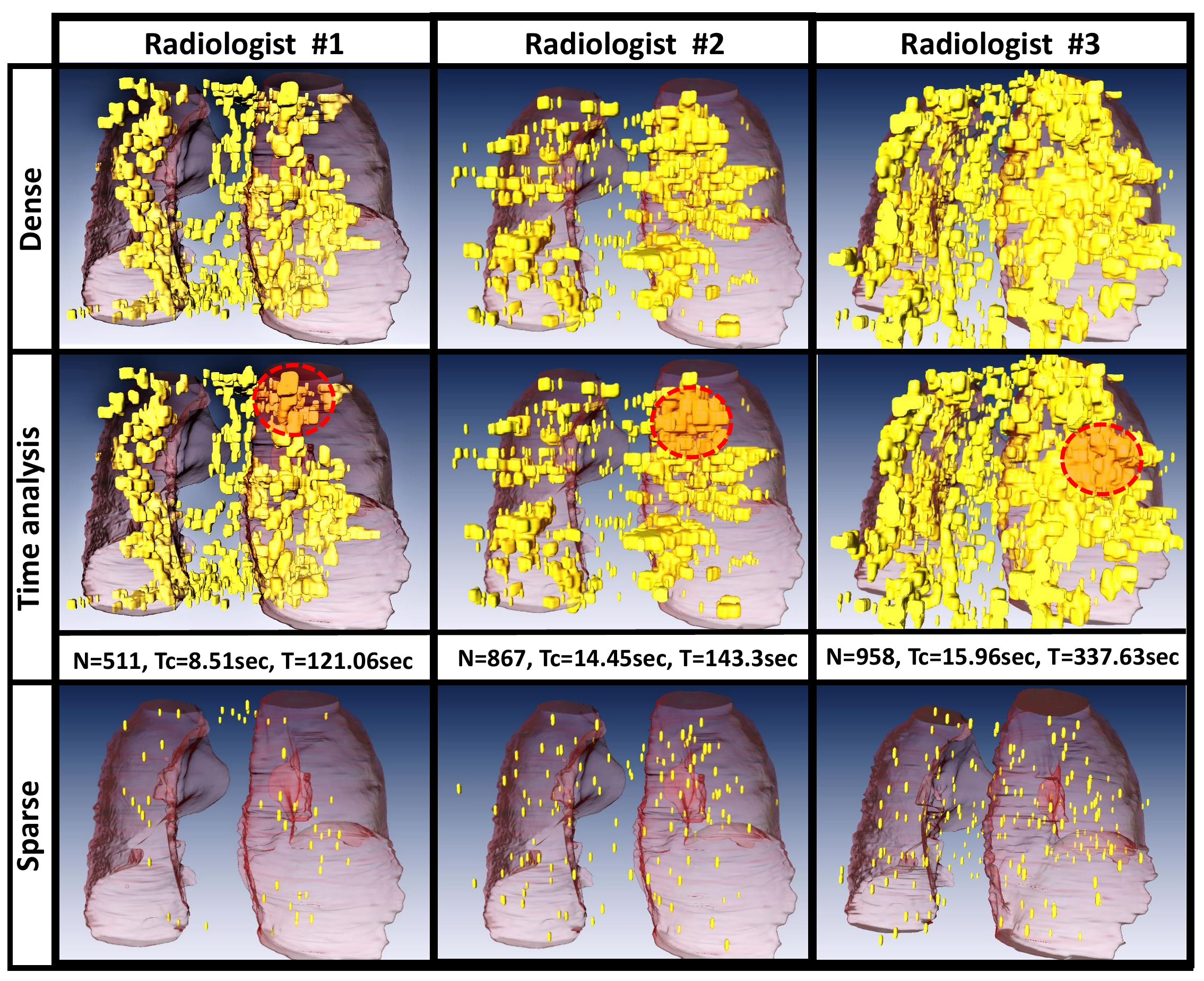}
\caption{Lung cancer screening experiment with CT data. Dense and sparse gaze points on 3D lung surface as well as time analysis. Number of nodes in the largest cluster (N), corresponding time spent by radiologist on that cluster (Tc) and overall screening time (T), with the eye-tracker frequency being $60Hz$, for each radiologist is computed. \label{fig:VRlung}}
\end{figure}

For a qualitative visualization on the image space (CT lungs), we showed the effect of our sparsification method on the dense eye-tracking data as well. Figure~\ref{fig:VRlung} shows the original gaze points, from $3$ radiologists, on the volume renderings of corresponding lung images in the first row. The second row illustrates the timing component of visual search patterns on the whole scan as well as the selected regions (i.e., attention region is indicated by circles) for each radiologist. The third row shows sampled data points after the clustering algorithm is applied. This figure supports how successful a very dense data can be sparsified for any local/global image analysis task easily.


\subsubsection{Quantitative results for sparsification:} 
\begin{figure}[h]
\centering
\includegraphics[scale=0.73]{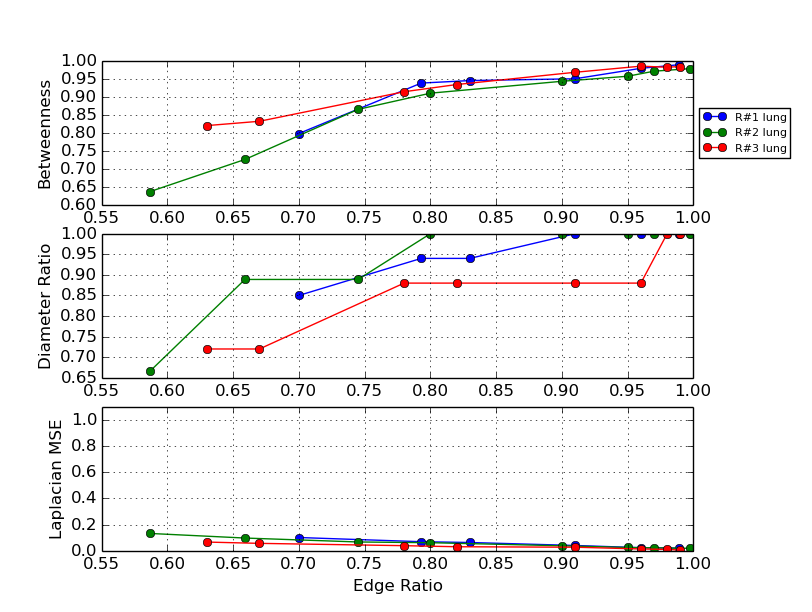}
\caption{Quantitative parameters to compare graph topology between already clustered data and sparsified  data with respect to the preserved edge ratio. $R\#$ indicates a particular radiologist (blue, green, red). (Lung cancer screening experiment)\label{fig:lungplot}}
\end{figure}

\begin{figure}[h]
\centering
\includegraphics[scale=0.73]{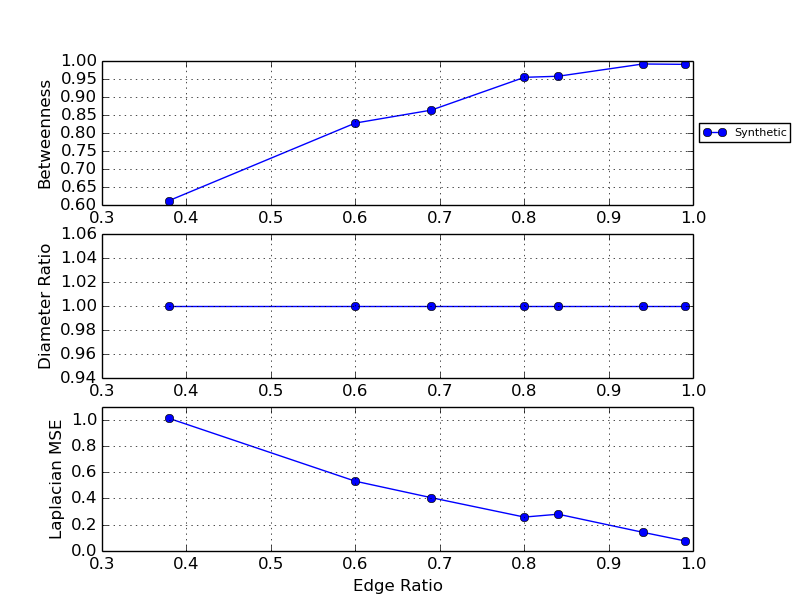}
\caption{Quantitative parameters to compare graph topology between already clustered and sparsified  data with respect to the preserved edge ratio. (Synthetic data experiment)\label{fig:syntheticplot}}
\end{figure}

In order to compare the topology of the graphs before and after sparsification, we reported the \textit{Diameter,} \textit{Betweenness,} and \textit{Mean Square Error (MSE)} of the graph Laplacian matrices. All these parameters are well established metrics representing structure of the graphs. \textit{Diameter:} measures the length of maximum shortest path in a graph, \textit{Betweenness:} is a measure of node centrality and counts the number of shortest paths that pass through a node. The Spearman's rank correlation coefficient is generally used to compare betweenness of the original and sparsified graph. \textit{MSE:} relates the structure of the graph before and after sparsification based on the error in the Laplacian matrix of the graph. The results for lung screening data and synthetic data are plotted in Fig. \ref{fig:lungplot} and Fig.~\ref{fig:syntheticplot}, respectively. The above-mentioned metrics for $3$ different radiologists are plotted in Fig. \ref{fig:lungplot}. Each point in the plot is computed corresponding to an edge ratio. The edge ratio is simply the ratio of edges in the graph after sparsification over the original graph. As expected, by removing more edges (edge ratio drops), betweenness and diameter metrics decrease and Laplacian MSE increases. 

\begin{figure}
\includegraphics[scale=0.8]{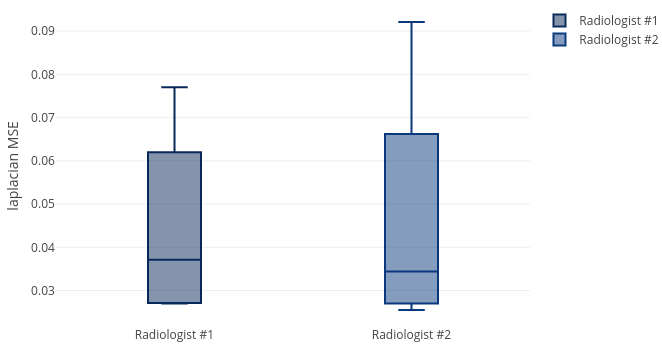}
\caption{Inter-observer variation of MSE for $2$ radiologists on $4$ different scans. \label{fig:lungvariationplot}}
\includegraphics[width=0.51\linewidth]{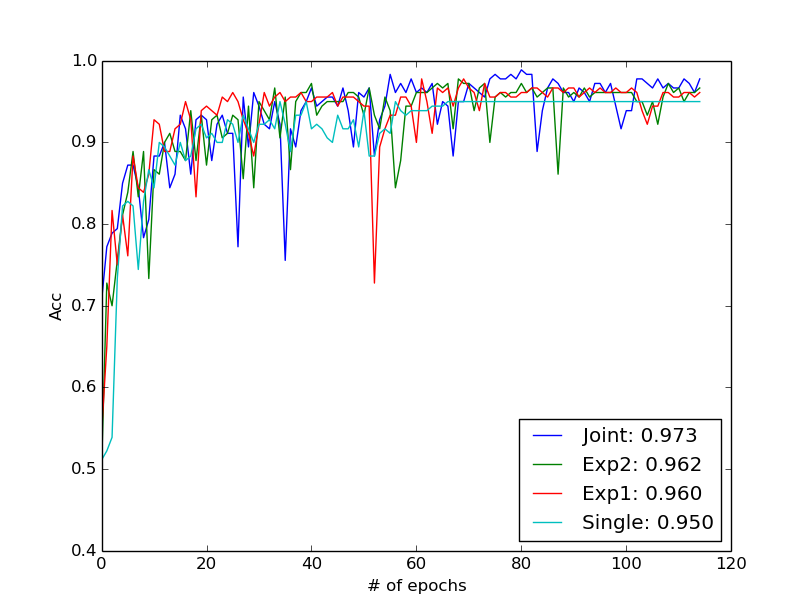}
 \includegraphics[width=0.51\linewidth]{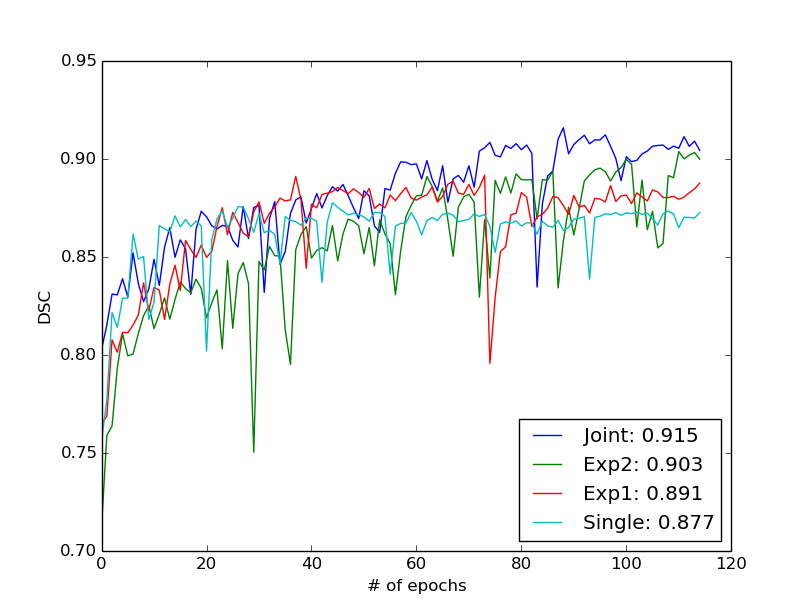}
  \caption{Left: Comparison of accuracy on the test set over training epochs. An increase from $95\%$ to $97\%$ is observed. Right: Comparison of DSC on test set is shown. An increase from $87\%$ to $91\%$ is observed in the network's trained state.}    \label{fig:jointvssingle}
\end{figure}

Variation of MSE for the data recorded from two radiologists (who read $4$ different scans) is plotted in Fig.~\ref{fig:lungvariationplot}. The MSEs are computed for the fixed edge ratio of $0.9$ in this analysis. With a fixed edge ratio, higher MSE means that the original graph is more sparse. This further indicates that removing the same ratio of edges distorts the graph structure and, leads to a higher MSE. Hence, the radiologists' pattern of search can be compared within this variation. A higher average MSE means that the radiologist is performing a targeted search and most probably is more expert radiologist.

\subsubsection{Evaluation of deep learning based diagnosis algorithm} 
The proposed MTL CNN achieved an average Dice Similarity Coefficient (DSC) of $91\%$ for segmentation and $97\%$ accuracy of classification (i.e., nodule vs. non-nodule). We also analyzed individual performances of segmentation and FP removal tasks, and their joint training (i.e., multi-task learning). For joint training, we designed two additional experiments to explore optimal network performance for the both tasks: 
\begin{itemize}
\item Exp1: We connected the fully connected layers to the encoder (at the network's bottle neck) and trained the network. In this experiment, classification loss only affected the encoder while segmentation loss affected the whole network.
\item Exp2: We trained the network by computing both loss function at the end of the network, similar to the joint training architecture, and obtained the features from the encoder (at the bottle neck) to train another fully connected layer and performed classification. In this experiment, both loss functions affected the whole network while classification is performed based on features extracted from the bottle neck.
\end{itemize}
As can be seen in Fig.~\ref{fig:jointvssingle}, both tasks benefited from joint training, and each outperformed the best of single task training. The plots indicate the DSC and classification accuracy over training epochs on validation sets (averaged). These experiments support the hypothesis that segmentation and FP reduction are highly correlated.

\begin{table}[h]
\centering
\caption{Comparison of Dice Similarity Coefficient (DSC) and classification accuracy of the proposed multi-task network with the baselines (single task networks) and methods from the literature corresponding to each individual task. }
\label{table:comparison}
\begin{tabular}{|l|c|l|c|l}
\cline{1-4}
\multicolumn{2}{|c|}{\cellcolor[HTML]{b2b2b2} \textbf{Segmentation}} & \multicolumn{2}{c|}{\cellcolor[HTML]{b2b2b2} \textbf{FP reduction}} &  \\ \cline{1-4}
\multicolumn{1}{|c|}{ \textbf{Method}} &  \textbf{DSC}  & \multicolumn{1}{c|}{ \textbf{Method}} &  \textbf{Accuracy} &  \\ \cline{1-4}
 \cite{khosravan2016gaze2segment}  &   $86.0\%$    &  \cite{dou2017multilevel}  &  $96.0\%$  &  \\ \cline{1-4}
 Single &   $87.7\%$  &    Single   &   $95.0\%$   &  \\ \cline{1-4}
 Exp1 &   $89.1\%$  &    Exp1   &   $96.0\%$   &  \\ \cline{1-4}
 Exp2 &   $90.3\%$  &    Exp2   &   $96.2\%$   &  \\ \cline{1-4}
   Joint  &    $\textbf{91.5\%}$    & Joint &  $\textbf{97.3\%}$   &  \\ \cline{1-4}
\end{tabular}
\end{table}

Table~\ref{table:comparison} shows the quantitative results of the proposed MTL-CNN. We compared the performance of our network with its own baselines, which are the \textit{single task} training of network on each task. For nodule segmentation and FP removal, we have selected two studies as example baselines for comparisons as well. Note that our proposed system is generic and each component of the network can be improved by replacing them with the most upto date successful segmentation and/or FP removal neural network counterparts as long as those alternative networks allow joint training for improved analysis.

\subsection{Feasibility study of multi-screen eye-tracking:}
As a proof-of-concept, we tested C-CAD on a multi-screen prostate MR screening experiment. Promising results show the flexibility and generalizability of our algorithm in dealing with more complex tasks including more than one screen.

This experiment was performed on a multi-parametric MRI scan of a single subject. MRI characteristics are: axial T2 weighted (T2w), with FOV of $140\times 140$ and resolution of $0.27\times 0.27\times 3~mm^{3}$, Dynamic Contrast Enhanced (DCE) with FOV of $262\times 262$ and resolution of $1.02\times 1.02\times 3$, $b=2000s/mm^{2}$, Diffusion Weighted Imaging (DWI) with FOV of $140\times 140$ and resolution of $0.55\times 0.55\times 2.73 mm^{3}$. Apparent Diffusion Coefficient (ADC) map was derived from 5 evenly spaced b value ($0-750s/mm^{2}$) DWI.

\begin{figure*}
\centering
\includegraphics[scale=0.4]{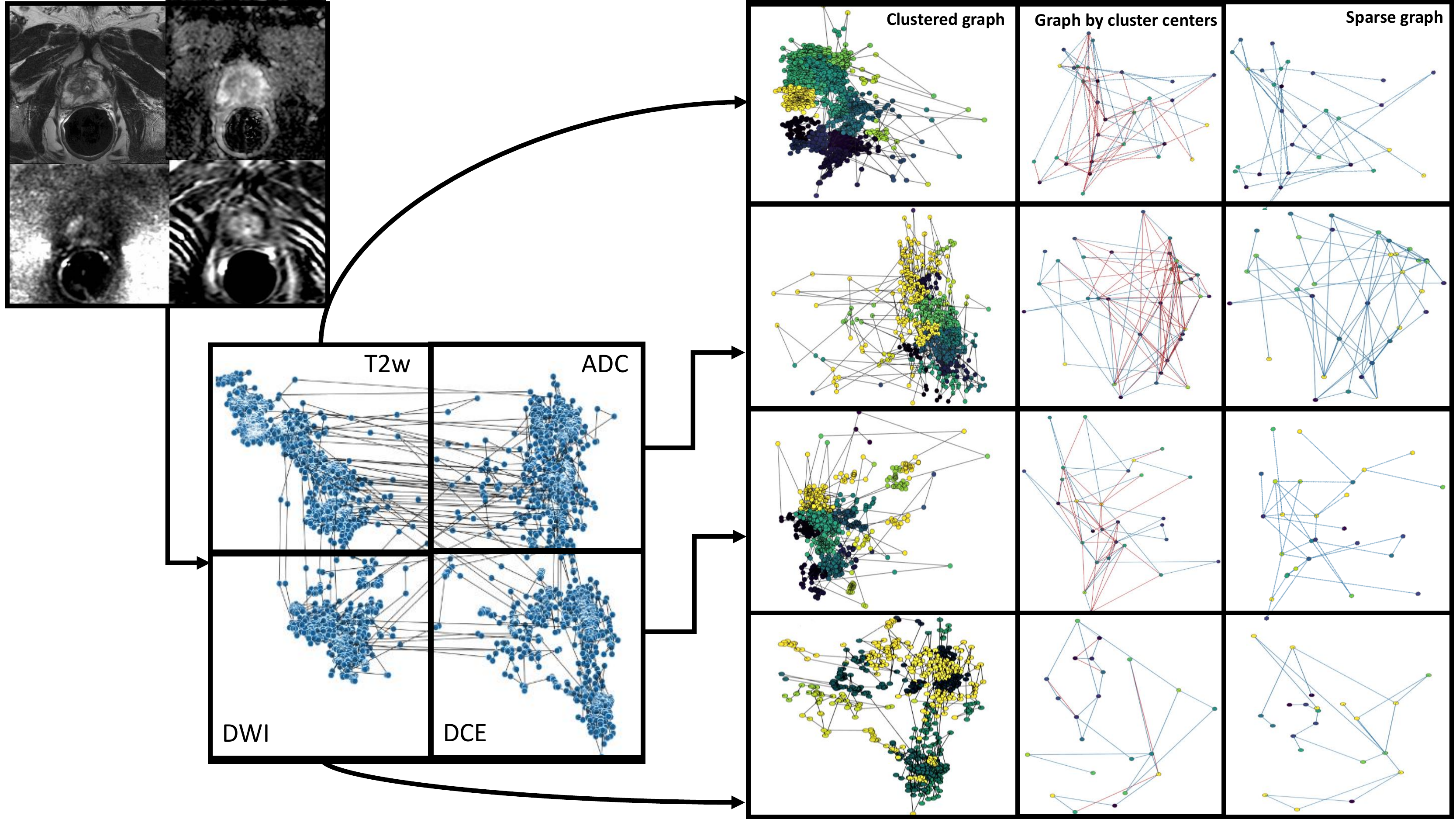}
\caption{Prostate cancer screening experiments with multi-parametric MRI. Left: four MRI modalities and corresponding dense gaze patterns. Right: Clustered and sparsified gaze patterns corresponding to each modality. First column: clustered dense gaze patterns. Second column: attention based clustering. Third column: sparse graph after further reducing edges. \label{fig:prostate}}
\includegraphics[scale=0.7]{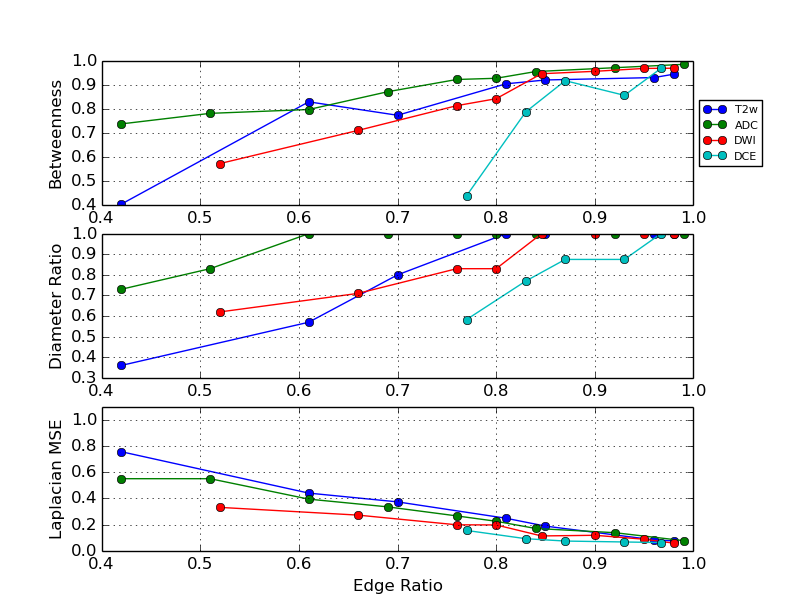}
\caption{Prostate screening experiment quantitative results.\label{fig:prostatequantitative}}
\end{figure*}

One of our participating radiologists, an expert in prostate cancer screening, examined multi-parametric MRI (four 3D images) for routine prostate cancer screening. Based on the results reported in Fig. \ref{fig:prostate}, it is evident by the sparsified graphs that the radiologist used axial T2-weighted images (anatomical information) and ADC maps (showing magnitude of diffusion) more frequently than other two images. This observation suggests that although all four modalities are being used for making a diagnostic decision T2-weighted and ADC map are more informative to the radiologists in the screening process. This observation can be useful in further developments of automatic computer-aided diagnosis systems.

\begin{figure}[h]
\centering
\includegraphics[scale=0.65]{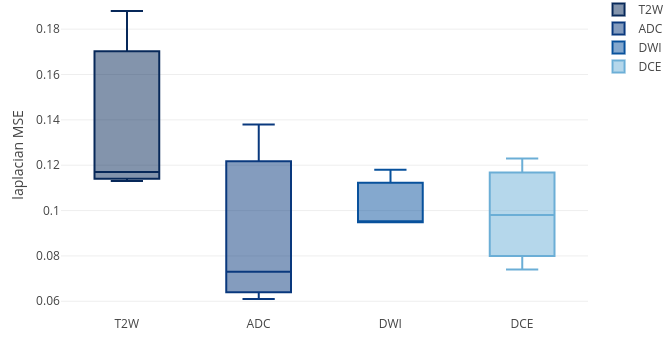}
\caption{Variation of MSE on different prostate images per modality.  \label{fig:prostatevariation}}
\end{figure}

Quantitative results of our method for different modalities as well as the variation over these modalities are shown in Fig. \ref{fig:prostatequantitative} and Fig. \ref{fig:prostatevariation}, respectively. In screening, DWI and DCE modalities were used less frequently than T2-weighted and ADC modalities; therefore, the initial graph representations of the DWI and DCE are less dense compared to those of T2-weighted and ADC. For those less dense graphs, the sparsification algorithm  achieved a similar MSE performance in most edge ratios larger than 0.5. From the reason that  the sparsification algorithm keeps the graph in a $\sigma-spectral~approximation$ of the original graph, we cannot remove large number of edges from less dense graphs. This situation is reflected in diameter ratio and betweenness metrics. 

\section{Discussions and concluding remarks}
Our study offers a new perspective on eye-tracking studies in radiology because of its seamless integration into the real radiology rooms and collaborative nature of image analysis methods. First, the proposed system models the raw gaze data from eye-tracker as a graph. Then, a novel attention based spectral graph sparsification method is proposed to extract global search pattern of radiologist as well as attention regions. Later, we propose a 3D deep multi-task learning based CNN to perform diagnosis and segmentation tasks jointly inside ROIs. Our proposed sparsification method reduced $90\%$ of data within seconds while keeping mean the square error under $0.1$. The segmentation algorithm achieved the average Dice Similarity Coefficient of $91\%$ and classification accuracy for nodule vs. non-nodule was $97\%$.

As can be interpreted from the lung screening experiment, the less experienced participant had more crowded visual search patterns and examined the most lung volume. Also, from the prostate cancer screening experiment, we observed that radiologists use anatomical/structural information more frequently than other modalities in screening (i.e., diffusion MRI). This potentially shows the importance of anatomical information in prostate cancer detection but at the same time we noticed that when anatomical information gives less clues to radiologists about potential abnormality, radiologists looks for complementary information from other imaging modalities to make inference. Our system provides visualization of this process for the first time in the literature. Scanpaths across different screens prove this observation as can be seen in Fig.~\ref{fig:prostate}. 

Our work has some limitations that should be noted. One of the limitations  is the lack of large amount of data for training more sophisticated deep learning models as well as conducting scanpath analysis with several other radiologists. In our extended study, we plan to address this limitation and explore the validity of the proposed methods in different settings, incorporating the behavioural patterns into screening experiments such as cognitive fatigue of the radiologists.

For lung cancer screening with C-CAD, our system has the assumption that the radiologists are examining only lung regions and the ROIs fall into the lung regions. If the radiologist starts focusing on some other areas, outside the lungs, the segmentation results might not be as desired, because of non-lung regions. To solve this problem, one may include a simple segmentation step into the C-CAD to restricts the ROI definition into the lungs only. However, this procedure may affect analysis of incidental findings too.  


In conclusion, CAD systems are often prone to high number of false positives findings, which is one of the main drawbacks in such systems. Missing tumors, especially in their early stages, is also very common in screening. To overcome these problems and increase the efficacy of the cancer screening process, we propose a novel computer algorithm, namely collaborative CAD (C-CAD). Our proposed method takes into account the gaze data of radiologists during the screening process, and incorporates this information into the CAD system for better accuracy in reducing false positive findings in particular. In return, C-CAD has the capability of improving true positive findings as well as reducing missing cases.
With our proposed attention based graph sparsification method, qualitative comparison and analysis of different radiologists' visual search patterns (both locally and globally) has become feasible. It is worth noting that we are doing a local image analysis in the regions of interest and not solving the detection problem per se. Since our framework is capable of integrating any image analysis block as well as a detection block, having a detection task on top of diagnosis and segmentation is a promising future direction which enables the framework to handle cases that are missed by radiologists. 


\section*{Acknowledgement }
We would like to thank Nancy Terry, NIH Library Editing Service, for reviewing the manuscript.
\section*{References}
\bibliography{mybibfile}

\end{document}